%% file: draft.tex
\documentclass[sigconf, natbib=false, nonacm]{acmart}

\usepackage[datamodel=acmdatamodel, style=acmnumeric, sorting=none, backend=biber]{biblatex}
\usepackage{amsmath, amsfonts}

\usepackage{orcidlink}

\usepackage{graphicx}
\usepackage{subcaption}

\usepackage[inline]{enumitem}
\setlist*[itemize]{labelindent=10pt, itemindent=0pt, leftmargin=*}

\usepackage{booktabs}
\usepackage{multirow}
\usepackage{tabularx}
\newcommand{\highlightG}[1]{\colorbox{green!80}{#1}}
\newcommand{\highlightY}[1]{\colorbox{yellow!80}{#1}}

\usepackage{pgfplots}
\pgfplotsset{compat=1.18}
\usepgfplotslibrary{statistics}

\usepackage[nameinlink]{cleveref}

\usepackage{csquotes}
\usepackage{textcomp}

\usepackage{balance}

\setcopyright{rightsretained}
\copyrightyear{2025}

\begin{document}

\title[InfoPos: A Design Support Framework for ML-Assisted FDI]{InfoPos: A Design Support Framework for ML-Assisted Fault Detection and Identification in Industrial Cyber-Physical Systems}

\author{Uraz {Odyurt}}
\orcid{0000-0003-1094-0234}
\affiliation{%
	\institution{Faculty of Engineering Technology, University of Twente}
	\city{Enschede}
	\country{The Netherlands}}
\email{u.odyurt@utwente.nl}

\author{Richard {Loendersloot}}
\orcid{0000-0002-1113-8203}
\affiliation{%
	\institution{Faculty of Engineering Technology, University of Twente}
	\city{Enschede}
	\country{The Netherlands}}
\email{r.loendersloot@utwente.nl}

\author{Tiedo {Tinga}}
\orcid{0000-0001-6600-5099}
\affiliation{%
	\institution{Faculty of Engineering Technology, University of Twente}
	\city{Enschede}
	\country{The Netherlands}}
\email{t.tinga@utwente.nl}

\renewcommand{\shortauthors}{U. Odyurt et al.}

\begin{abstract}
The variety of building blocks and algorithms incorporated in data-centric and ML-assisted fault detection and identification solutions is high, contributing to two challenges: selection of the most effective set and order of building blocks, as well as achieving such a selection with minimum cost. Considering that ML-assisted solution design is influenced by the extent of available data and the extent of available knowledge of the target system, it is advantageous to be able to select effective and matching building blocks. We introduce the first iteration of our InfoPos framework, allowing the placement of fault detection/identification use-cases based on the available levels (positions), i.e., from poor to rich, of knowledge and data dimensions. With that input, designers and developers can reveal the most effective corresponding choice(s), streamlining the solution design process. The results from a demonstrator, a fault identification use-case for industrial Cyber-Physical Systems, reflects achieved effects when different building blocks are used throughout knowledge and data positions. The achieved ML model performance is considered as the indicator for a better solution. The data processing code and composed datasets are publicly available.

\end{abstract}

%


\maketitle

\input{text/body}

\balance

\printbibliography

\end{document}

%% file: text/body.tex

\section{Introduction}
\label{sec:introduction}
One of the important activities involved in a successful strategy towards predictive maintenance for industrial Cyber-Physical Systems (CPS) is anomaly detection and identification. In the context of industrial CPS, data-centric solutions consuming time-series data from machine sensors, have proven to be highly capable~\cite{Odyurt:2022:IRIC}. During the design phase of such solutions, there are numerous data processing and Machine Learning (ML) algorithms, suitable for time-series data analysis, to choose from. Generally speaking, with industrial CPS, data is available in abundance, which can be collected from a multitude of available sensors. Typically, these machines are intended to operate non-stop and at full capacity, requiring any data collection and monitoring to be well-planned.

Contrary to one's initial assumption, the abundance of data could present a challenge. Besides the complexities and resource costs imposed by excessive data collection, large amounts of data do not necessarily lead to better predictions, for example, when there is a disproportion between normal and anomalous data. As such, \emph{it is highly advantageous to be able to select the right data processing steps, choose the most suitable ML algorithm, and focus on the most effective portion of the data}.

It is even more advantageous to know which of the above ingredients (data processing, ML algorithm, and data subset) match and work best together, enabling the selection of the most effective combination, should one component be constrained. For instance, if available data is limited to a specific portion, the best complementary ML algorithm should be considered. Most importantly, all such compatibilities should preferably be known upfront, i.e., data-awareness.

\begin{figure}[htbp]
	\centering
	\includegraphics[width=0.9\linewidth]{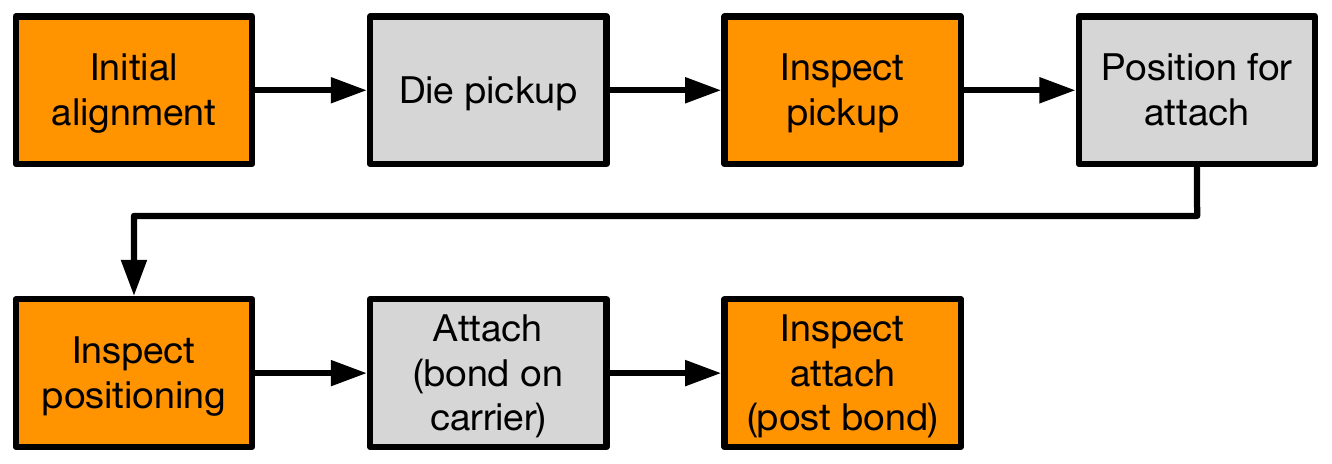}
	\caption{An example CPS machine cycle, involving action steps (in grey) and inspection steps (in orange).}
	\label{fig:machine_cycle}
\end{figure}

\paragraph*{Scope of use-cases}
Our domain of applicable use-cases is strictly industrial CPS. Examples include semiconductor photolithography machines, production printing machines, die bonder machines, and similar equipment. These systems share several characteristics: presence of highly complex, multi-node compute and control elements; a narrowly defined domain of operational tasks, i.e., highly purpose-built; highly repetitive and predominantly sequential machine cycles; constrained yet non-deterministic behaviour; and a continuous focus on achieving high-yield production output. Take an ITEC\footnote{\url{https://www.itecequipment.com/}} die bonder machine for instance. It is a repetitive high-yield industrial CPS, dedicated to the task of die bonding. These machines are controlled by computing nodes, real-time and non-real-time, with sequential machine cycles per die. A high-level diagram of such machine cycles is drawn in \Cref{fig:machine_cycle}.

\paragraph*{Contribution}
We introduce the first iteration of our \emph{InfoPos framework}, intended to support designers and engineers in the selection of most effective elements when building ML-assisted Fault Detection and Identification (FDI) solutions for industrial CPS. Examples of such element variations are the type of ML algorithm, data processing/transformation steps applied, or the level of these steps, and the considered portion of data. In short, we provide:
\begin{itemize}
	\item The InfoPos framework as an analysis/design support tool for ML-assisted fault identification design composition and tuning.
	\item Preliminary results from a real-world platform as a demonstrator use-case, covering numerous combinations of available knowledge, available data and traditional ML algorithms. Our primary focus for this paper is the knowledge dimension and fault identification using traditional ML. The data dimension and deep neural networks will be covered in future work.
	\item Publicly available processed datasets~\cite{Odyurt:2025:DATASET} and the data workflow code~\cite{Odyurt:2025:CODE}, covering the data processing and ML model training.
\end{itemize}

\paragraph*{InfoPos usage}
The application of InfoPos as a conceptual framework is twofold. When faced with many options and potential combinations for a fault identification solution intended for industrial CPS, InfoPos allows for a structured and systematic analysis of possibilities, assisting with elimination of undesirable cases. Secondly, InfoPos assists designers with selection or narrowing down of solution variables. Based on our results, being aware of the knowledge position upfront will point out the most effective combination of elements. The fault identification solution using traditional ML included in this paper has shown different performance levels in the presence of varying internal system details. Applying the same solution to new use-cases, designers will be aware of the level of available knowledge and tweak the solution accordingly. In other words, the possibility of applying informed segmentation will be known upfront. Following our framework will drastically reduce search and optimisation effort for a FDI solution.

After this introduction, definitions for the important concepts are provided in \Cref{sec:background}. Our methodology for both the anomaly identification solution and the formulation of different positions for data and knowledge is described in \Cref{sec:methodology}. Results from the application of anomaly identification in different positions is given in \Cref{sec:results}, alongside discussions on these results. \Cref{sec:related_work} covering a compact related work is followed by \Cref{sec:conclusion} with concluding remarks.

\section{Background and definitions}
\label{sec:background}
To explain the perspective and what the roles of knowledge and data are in shaping data-centric and ML-assisted solutions, it is important to clarify the terminology first. Throughout this paper, what is considered as \emph{data} is primarily metric traces collected from a multitude of available sensors, a.k.a., Extra-Functional Behaviour metrics. Modern industrial CPS machines are equipped with a variety of sensors, mainly intended for product quality control. Both individual hardware sensors, e.g., a torque measuring sensor, a voltage collector, or a temperature sensor, as well as software sensors are considered. The latter refers to system resource monitoring via virtual metric collectors to record variables such as computational time, memory usage and so forth. This type of sensing is applicable to the compute and control elements.

\emph{Knowledge} can be sourced from different artefacts, e.g., blueprints, system/machine logs (not to be confused with traces), design documentation, and expert knowledge. System knowledge reveals its operational sequence(s), functional and data dependencies, applied configuration, input material parameters, physical environment specifics, and possibly reference benchmarks. For example, size and type of input, production rate (which could be translated to frequency or required yield), machine cycle steps and their order, are all parts of this knowledge. Note that techniques surrounding the encapsulation of knowledge, i.e., knowledge representation, are out of the scope of this paper.

It is important to emphasise that industrial CPS, as stated in \Cref{sec:introduction}, refer to purpose-built and repetitive complex systems, i.e., machinery intended to excel at a very limited domain of operational tasks, at high rates and with high yields. These complex systems consist of numerous dedicated subsystems, expected to work cohesively.

\subsection{Knowledge and data}
\label{subsec:knowledge_and_data}
The two major dimensions influencing the design and the effectiveness of ML-assisted solutions data processing solutions, are the \emph{knowledge position} and the \emph{data position}. In this context, the knowledge position refers to the present level of understanding for the system's internals, its interactions with the physical domain, and how it is related to any accompanying data. Similarly, the data position refers to how extensive, complete, and granular the collected or available data is. This could point to qualities such as descriptiveness, comprehensiveness, and accuracy\footnote{Accuracy refers to the absence/presence/amount of noise.} of collected data, in relation with the represented complex system.

To avoid a circular definition for rich and poor data, the highest quality collectible data is always considered as the rich position. This will depend on the system at hand, its included sensors, as well as the monitoring and data collection subsystems. In short, the data position spectrum is system-specific. Based on our interactions with industrial stakeholders, the extent of achievable data richness is a known factor, i.e., richest data position is always known. As such, poor data positions, which could be encountered at production, shall be defined relative to such a reference.

\begin{figure}[htbp]
    \centering
    \begin{subfigure}{\linewidth}
    	\centering
	    \includegraphics[width=0.7\linewidth]{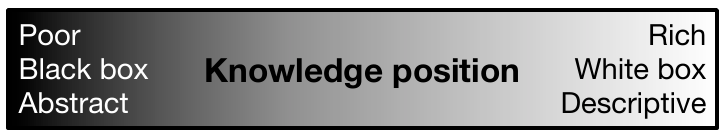}
	    \caption{Knowledge spectrum with representative extremities.}
	    \label{fig:knowledge_spectrum}
    \end{subfigure}
    \qquad
    \begin{subfigure}{\linewidth}
    	\centering
    	\includegraphics[width=0.7\linewidth]{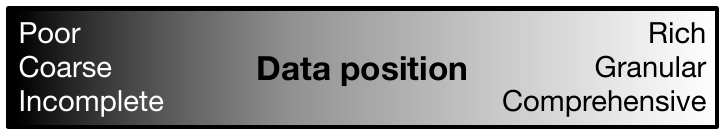}
		\caption{Data spectrum with representative extremities.}
		\label{fig:data_spectrum}
    \end{subfigure}
	\caption{Knowledge and data positions as the two main dimensions affecting data-centric solutions.}
	\label{fig:spectrums}
\end{figure}

Both dimensions are to be considered as a spectrum, spanning from a poor state to a rich one. \Cref{fig:spectrums} provides examples of opposing states for knowledge and as depicted in \Cref{fig:knowledge_spectrum}, abstract and black-box versus descriptive and white-box representations come to mind. For data, as shown in \Cref{fig:data_spectrum}, coarse or incomplete versus granular or comprehensive data is plausible.

\subsection{Information positions}
\begin{figure}[htbp]
	\centering
	\includegraphics[width=0.8\linewidth]{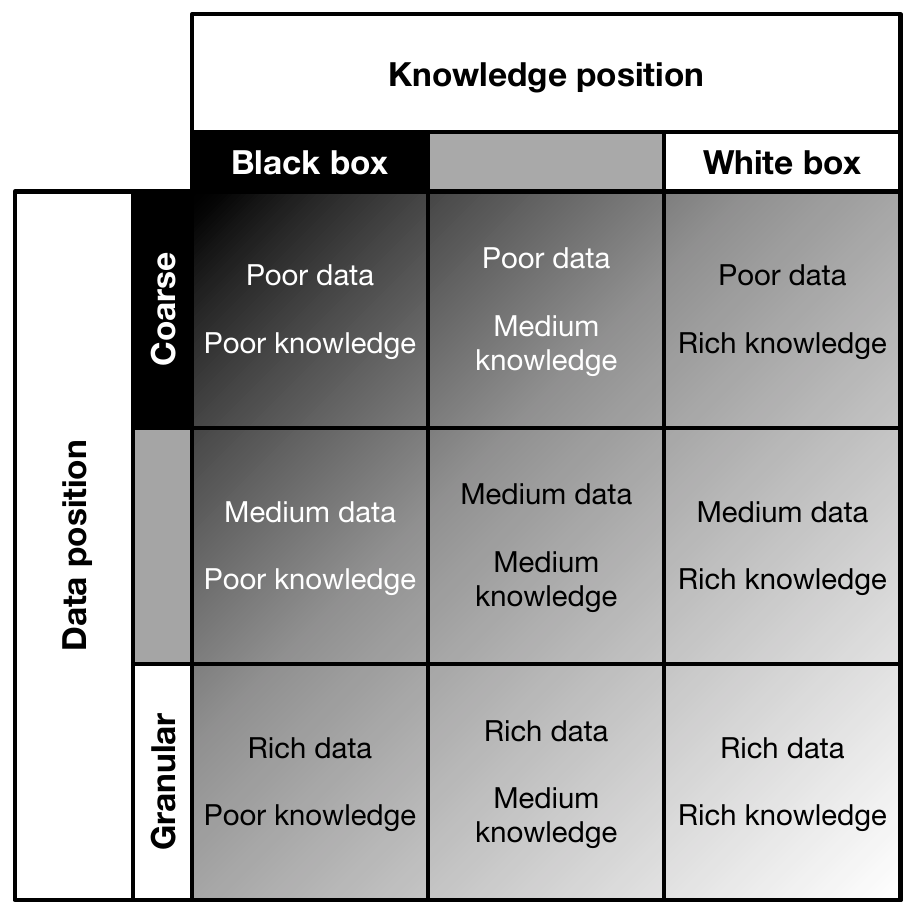}
	\caption{Information position matrix resulting from the composition of knowledge and data dimensions.}
	\label{fig:infopos_quadrant}
\end{figure}

With both dimensions taken into account, any solution design task could land on either of the cells from the $3 \times 3$ matrix given in \Cref{fig:infopos_quadrant}.

Depending on practical circumstances involved with the use-case at hand, one can expand or shrink the quadrant by adding or removing steps to/from each dimension. To simplify the demonstration and to deliver the message, only considering the very extreme cases is a suitable approach.

\section{Methodology}
\label{sec:methodology}
A demonstrator platform from~\cite{Odyurt:2021:PPFT} and the associated data collected from it is the source. The main advantage of this platform is the collection of real and balanced data, i.e., not synthetic. Though the scale of the platform is small, it reflects the real-world task of continuous live image processing. Image analysis using a pre-trained ML model is performed as a computational workload (not to be mistaken with ML models used in our anomaly identification flow) to detect the presence of specific objects.

\begin{figure}[htbp]
	\centering
	\includegraphics[width=0.9\linewidth]{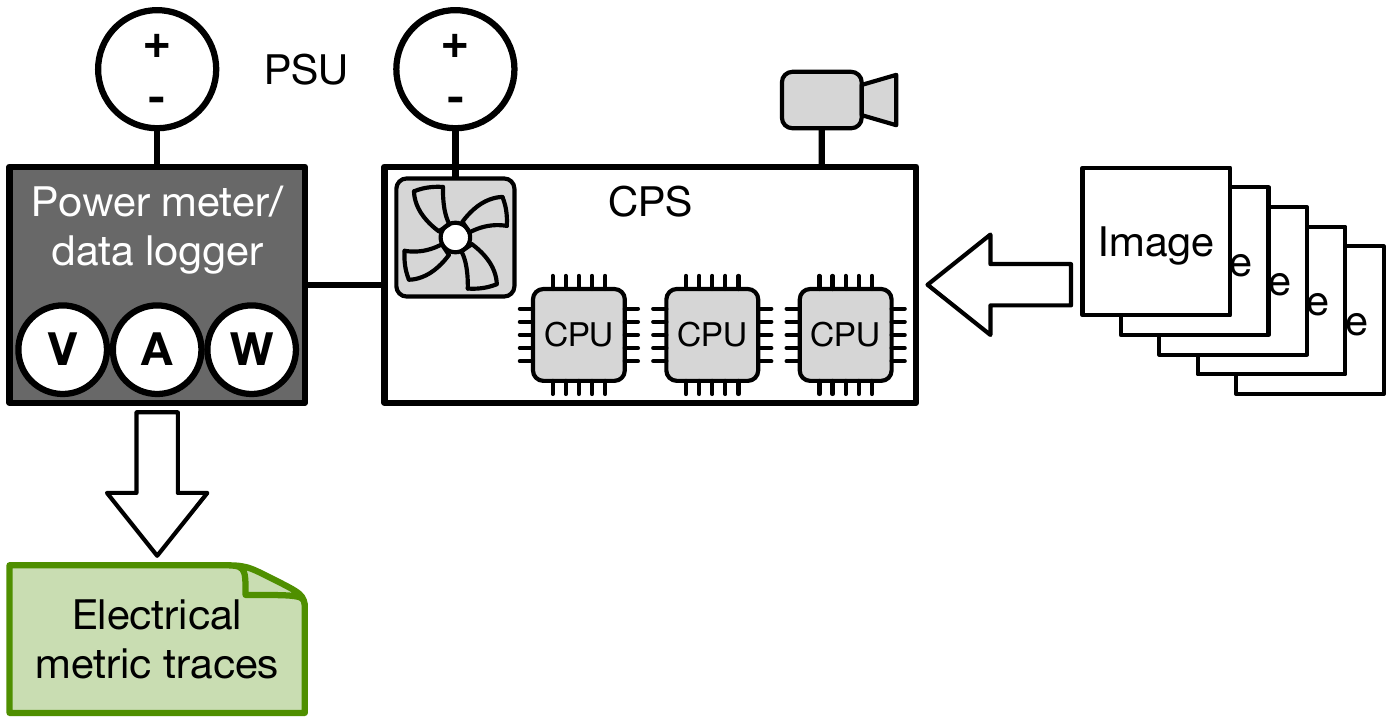}
	\caption{Data collection from the demonstrator set-up, including a dedicated electrical data logger.}
	\label{fig:demonstrator_setup}
\end{figure}

The data collection experimental set-up is covered in \Cref{fig:demonstrator_setup}, with the presence of a dedicated power data logger with an isolated power supply to avoid interference. This demonstrator might seem simplistic at first glance. However, considering the scope of purpose-built and repetitive industrial CPS, the likes of ITEC die bonders from \Cref{sec:introduction}, going down the subsystem hierarchy of real-world cases often leads to very similar execution timelines and subsystem cycles. In essence, considered source data mimics subsystem-level behaviour of an industrial CPS rather well.

The metric data is collected systematically during sequential operation of the demonstrator using varying inputs (2 batches of images), varying hardware configuration (2 available processing core types), varying repetition counts (2) and varying executional conditions. These conditions follow normal execution, execution under cooling subsystem failure, and execution under fluctuating power delivery, leading to 3 labels: Normal, NoFan, and UnderVolt, respectively.

\subsection{Data processing workflow}
\begin{figure*}[htbp]
	\centering
	\includegraphics[width=0.9\textwidth]{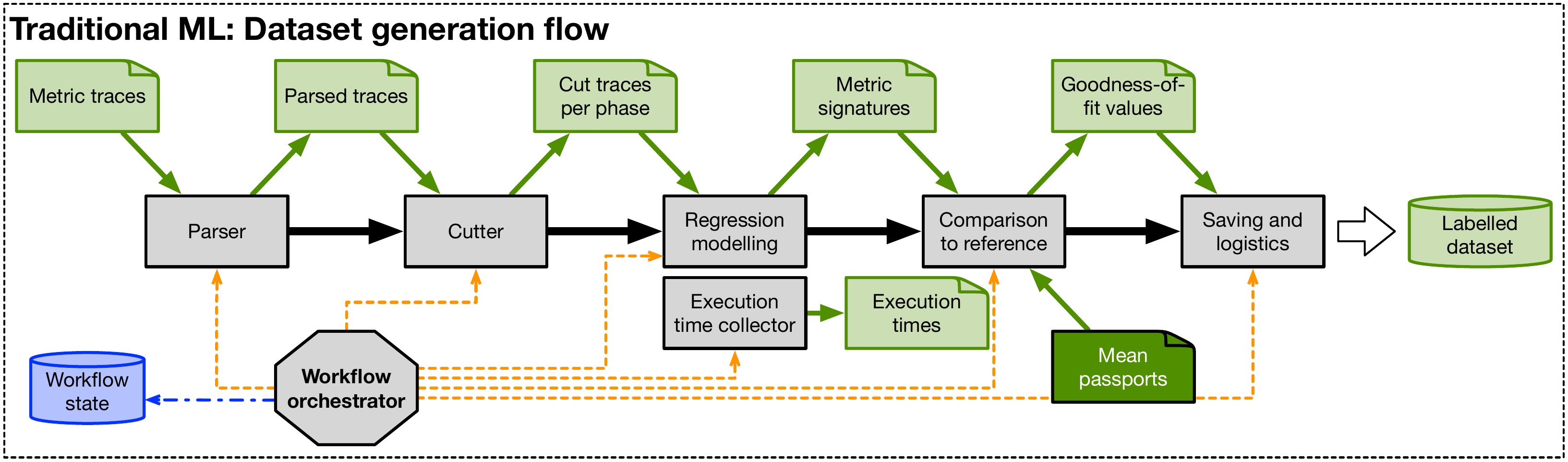}
	\caption{Our detailed data processing workflow, covering different steps, as well as the in-house simple orchestrator to run the workflow in parallel and at scale.}
	\label{fig:data_processing}
\end{figure*}

The preprocessing applied to the collected electrical metrics\footnote{Voltage is collected, but not considered.}, i.e., \emph{current}, \emph{power} and \emph{energy}, is depicted in \Cref{fig:data_processing}. Note that a similar preceding workflow generated the Mean Passport information, which will act as the reference point for comparing unknown execution data. Mean Passports are signatures belonging to executions with no anomalies, i.e., normal behaviour (labelled as Normal). These signatures are regression functions generated from data segments (phases) and could be in any degrees, linear, quadratic and more.

The extensive nature of preprocessing is intended to generate features required for traditional ML algorithms, a strategy that has proven rather effective. This solution~\cite{Odyurt:2022:IRIC} shall not be described further, as the focus of this paper is to evaluate the outcome of the solution under varying knowledge positions.

\subsection{Dataset}
The final output from the preprocessing workflow is a labelled dataset used for supervised ML model training and testing. Included feature columns are:
\begin{itemize}
	\item The time span covered by the data segment, i.e., the cut trace (\texttt{execution\_time}).
    \item Different parameters from linear or quadratic regression functions, representing the data segment \\(\texttt{coefficient\_2}, \texttt{coefficient\_1}, \texttt{intercept}).
    \item Different goodness-of-fit comparison calculations, quantifying the diversion of the unknown execution data from the reference execution data \\(\texttt{R2}, \texttt{R2\_absolute\_diff}, \texttt{RMSE}, \texttt{RMSE\_absolute\_diff}).
\end{itemize}

Considering the 8 data collection cases described in~\cite{Odyurt:2021:PPFT} (pointed earlier in this section, 2 inputs $\times$ 2 core types $\times$ 2 repetition counts), as well as the three experiment conditions applied, i.e., Normal, NoFan, and UnderVolt, we end up with 24 data collection scenarios. For each scenario, three quartile-based phase cuts (reductions or segmentations), alongside the full phase data (see \Cref{fig:uninformed_segmentation}) are considered. As such, there will be 4 phase data cuts per scenario, i.e., \emph{ini}, \emph{mid}, \emph{end}, and \emph{full}, resulting in 96 individual cases to be processed by this solution workflow. 
Needless to say, it is trivial to combine such data, as the format and headers are the same in all. These datasets are applied separately during ML model training and provide relevant results in separate tables in \Cref{sec:results}.

\subsection{Data segmentation}
\begin{figure}[htbp]
    \centering
    \begin{subfigure}{\linewidth}
    	\centering
	    \includegraphics[width=\linewidth]{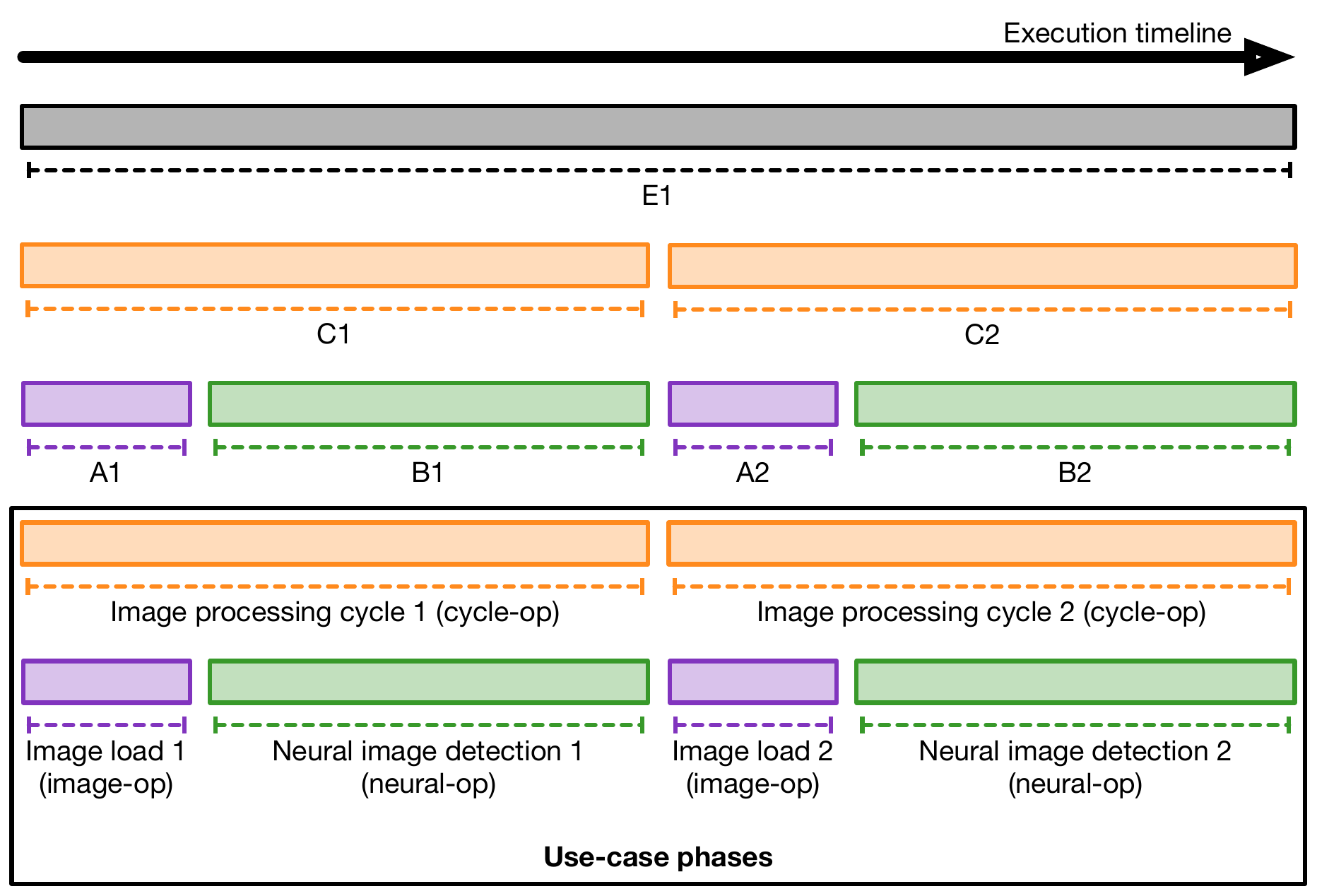}
	    \caption{Informed segmentation}
	    \label{fig:informed_segmentation}
    \end{subfigure}
    \qquad
    \begin{subfigure}{\linewidth}
    	\centering
    	\includegraphics[width=\linewidth]{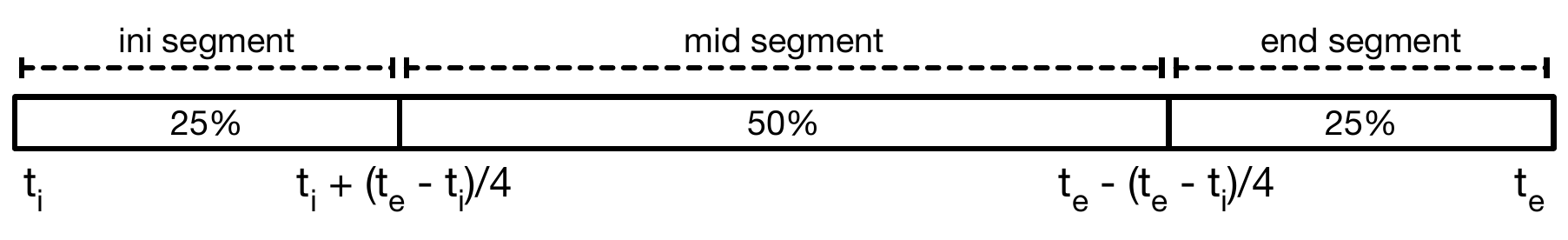}
		\caption{Uninformed segmentation}
		\label{fig:uninformed_segmentation}
    \end{subfigure}
	\caption{Different types of segmentation depending on the availability of the operational knowledge. Note the bottom two rows in (a), which correspond to phases representing tasks and sub-tasks from our demonstrator use-case.}
	\label{fig:data_segmentation}
\end{figure}

One of the steps most dependent on the available knowledge is segmentation (cutting) of data. There can be two segmentation types, informed, which cuts the data into known phases, or uninformed, which is independent of the internal operational knowledge of the system, resulting in a simple segmentation. Both types are depicted in \Cref{fig:data_segmentation}.

\paragraph*{Phase-based (informed) segmentation}
Phase-based segmentation is the informed type of segmentation. In the considered use-case, images processing is the computational workload. As any processing, this activity is not done in one step. The processing of a single data instance (an image) is covered by the \texttt{cycle-op} phase type, hence, one cycle of operation for this platform. Each cycle is composed of two inner and sequential phase types, \texttt{image-op} and \texttt{neural-op}, to load the image and to apply inference, respectively. The knowledge of this design and the knowledge of start and end events per phase type allows us to cut the metric data into chunks associated with each phase type. In \Cref{fig:informed_segmentation}, C1 can be considered as a \texttt{cycle-op} phase, composed of A1 and B1 corresponding to \texttt{image-op} and \texttt{neural-op} phases. Without the knowledge of internals (a poor knowledge position), only the data for \texttt{cycle-op} phases can be encapsulated.

\paragraph*{Quartile-based (uninformed) segmentation}
Independent from such knowledge, segmentation of data based on phase execution time quartiles can be considered. This is a rather simple, but effective segmentation strategy. Basically, any phase type's execution duration can be divided into 4 quartiles. Data contained in the first and the last are considered as \emph{ini} and \emph{end} segments, while the data from the two middle quartiles is the \emph{mid} segment, as shown in \Cref{fig:uninformed_segmentation}. It is important to note that, as a general rule, quartile-based segmentation can be applied to any phase, defined with or without internal knowledge. \emph{The motivation behind quartile-based segmentation lies in the presence of cold-start at the beginning and comparable effects at the end of most computational tasks}. At the same time, it is fair to assume that \emph{ini} and \emph{end} segments could be a better source for rare faults, specific to such regions.

\subsection{Simulating data positions}
\label{subsec:simulating_data_positions}
As mentioned in \Cref{subsec:knowledge_and_data}, the richest position for data is limited by the data collection capability of the monitoring subsystem. As we possess this reference data, to be able to experiment with a variety of data positions from rich to poor, lower quality data can be systematically generated through the application of algorithmic degradation techniques, e.g., Jitter, Time warping, Time masking, Amplitude scaling, Spike injection, Decimation (down-sampling). To arrive at different data positions, one can apply any or a combination of these techniques, at different rates, to effectively simulate lower data qualities. The goal is not to invoke system instability, but to achieve poor data representation of the system, at different levels. These degradations are applied to the \enquote{Cut traces per phase} intermediate artefact from \Cref{fig:data_processing}, to be followed by the rest of the data processing steps.

\subsection{ML algorithms for fault identification}
An exhaustive collection of traditional ML algorithms are considered in our experiments. These algorithms are, Boosted Decision Tree (BDT)~\cite{Friedman:2001:BDT}, Decision Tree (DT)~\cite{Breiman:1984:DT}, Extra Trees (ET)~\cite{Geurts:2006:ET}, Gaussian Naive Bayes (NB), Kernel Support Vector Machine (SVM), Linear Support Vector Classification (SVC) and Random Forest (RF)~\cite{Breiman:2001:RF}. These model types are utilised as multi-class classifiers and identify the type of system behaviour. The Normal behaviour, as well as two anomalous behaviours (NoFan and UnderVolt) are covered in experiments. The training is supervised and the list of classes can be easily expanded if representative data exists. Both prediction accuracy and F1 score are considered for model performance evaluation.

For our training, 3-fold cross-validation is applied and the average accuracy and average F1 score from all folds are calculated. In each experiment, models are trained with specific portions of data, resulting from aforementioned segmentation strategies. Note that while the best model performance is sought after, the primary goal is to discover the interplay between different scenario variables, making up the information position for that particular scenario.

\section{Results}
\label{sec:results}
Considering the high number of cases, variety of metrics and the number of ML algorithms, we end up with extensive results, of which we only discuss the most interesting bit. We have seen in previous research~\cite{Odyurt:2021:PPFT} and repeated the same observation that the most effective metric to consider in these experiments is \emph{electrical current}, leading to highest classification performances. This is valid throughout.

Considering that the datasets are well-balanced, prediction and F1 score calculations match rather well and either one can be considered as a single indicator of model performance. We do provide both, but rely on model accuracy to draw conclusions, which is corroborated by the F1 score as well.

Tree-based algorithms excel at this type of classification. Tree-based traditional ML refers to algorithms using decision trees or ensembles of decision tree. As such, we only focus on and discuss the results from BDT, DT, ET and RF classifiers.

Detailed results provided in \Cref{tab:model_performance} cover model performance metrics for the aforementioned classifiers, for numerous data segments. In particular, results dedicated to each data cut with uninformed segmentation, i.e., \emph{full}, \emph{ini}, \emph{mid} and \emph{end}, are provided separately in \Cref{tab:model_performance_full,tab:model_performance_ini,tab:model_performance_mid,tab:model_performance_end}, respectively. Here, the \emph{full} type is the representation of complete data. All available phase types, as well as their combinations as input for the ML model training are covered. For instance, phase type \enquote{all} refers to the use of data from all three individual phase types, i.e., \texttt{cycle-op}, \texttt{image-op}, and \texttt{neural-op}. Note that the three phase types are the result of informed segmentation, utilising the knowledge from system's internal operation. A number of immediate implications that can be observed are discussed in the following sections.
\begin{table*}[htbp]
    \centering
    \caption{Model performance results for training datasets composed in different knowledge positions are provided. Tables are organised per different uninformed segment cut, with: (a) full-cut, (b) ini-cut, (c) mid-cut, and (d) end-cut. Top classification results for richer and poorer knowledge positions are indicated with green and yellow highlights, respectively.}
    \label{tab:model_performance}
    \begin{subtable}{\textwidth}
        \centering
        \caption{Model performance results for full-cut segmentation, i.e., no segmentation, applied to each phase type}
        \label{tab:model_performance_full}
	    \begin{tabular}{@{}llrrrrrrrr@{}}
	        \toprule
	        \multicolumn{1}{c}{\textbf{Phase type}} & 
	        \multicolumn{1}{c}{\textbf{Signature}} & 
	        \multicolumn{1}{c}{\textbf{BDT acc.}} & 
	        \multicolumn{1}{c}{\textbf{BDT F1}} & 
	        \multicolumn{1}{c}{\textbf{DT acc.}} &
	        \multicolumn{1}{c}{\textbf{DT F1}} &
	        \multicolumn{1}{c}{\textbf{ET acc.}} &
	        \multicolumn{1}{c}{\textbf{ET F1}} &
	        \multicolumn{1}{c}{\textbf{RF acc.}} &
	        \multicolumn{1}{c}{\textbf{RF F1}} \\
	        \midrule
	        all						& Linear reg.	& 95.71\%  & 0.96  & 95.83\%  & 0.96  & 95.99\%  & 0.96  & 96.27\%  & 0.96 \\
	        cycle-op 				& Linear reg.	& 98.88\%  & 0.99  & 98.40\%  & 0.98  & 98.78\%  & 0.99  & 98.91\%  & 0.99 \\
	        image-op 				& Linear reg.	& 91.44\%  & 0.91  & 89.96\%  & 0.90  & 91.64\%  & 0.92  & 91.90\%  & 0.92 \\
	        neural-op 				& Linear reg.	& \highlightG{99.19\%}  & 0.99  & \highlightG{99.14\%}  & 0.99  & 98.93\%  & 0.99  & \highlightG{99.11\%}  & 0.99 \\
	        image-op + neural-op 	& Linear reg.	& 94.75\%  & 0.95  & 94.22\%  & 0.94  & 95.12\%  & 0.95  & 95.30\%  & 0.95 \\
	        \midrule
	        all 					& Quadratic reg.	& 96.16\%  & 0.96  & 95.94\%  & 0.96  & 96.44\%  & 0.96  & 96.60\%  & 0.97 \\
	        cycle-op 				& Quadratic reg.	& \highlightY{99.03\%}  & 0.99  & 98.78\%  & 0.99  & \highlightY{99.06\%}  & 0.99  & 98.93\%  & 0.99 \\
	        image-op 				& Quadratic reg.	& 92.15\%  & 0.92  & 89.89\%  & 0.90  & 92.81\%  & 0.93  & 92.81\%  & 0.93 \\
	        neural-op 				& Quadratic reg.	& \highlightG{99.21\%}  & 0.99  & 98.76\%  & 0.99  & \highlightG{99.11\%}  & 0.99  & \highlightG{99.06\%}  & 0.99 \\
	        image-op + neural-op 	& Quadratic reg.	& 95.16\%  & 0.95  & 94.49\%  & 0.94  & 95.80\%  & 0.96  & 95.80\%  & 0.96 \\
	        \bottomrule
		\end{tabular}
	\end{subtable}
    \vspace{1em}
	\begin{subtable}{\textwidth}
        \centering
        \caption{Model performance results for ini-cut segmentation, applied to each phase type}
        \label{tab:model_performance_ini}
        \begin{tabular}{@{}llrrrrrrrr@{}}
            \toprule
            \multicolumn{1}{c}{\textbf{Phase type}} & 
            \multicolumn{1}{c}{\textbf{Signature}} & 
            \multicolumn{1}{c}{\textbf{BDT acc.}} & 
            \multicolumn{1}{c}{\textbf{BDT F1}} & 
            \multicolumn{1}{c}{\textbf{DT acc.}} &
            \multicolumn{1}{c}{\textbf{DT F1}} &
            \multicolumn{1}{c}{\textbf{ET acc.}} &
            \multicolumn{1}{c}{\textbf{ET F1}} &
            \multicolumn{1}{c}{\textbf{RF acc.}} &
            \multicolumn{1}{c}{\textbf{RF F1}} \\
            \midrule
            all               		& Linear reg.	& 93.67\%  & 0.94  & 93.12\%  & 0.93  & 93.85\%  & 0.94  & 94.10\%  & 0.94 \\
            cycle-op          		& Linear reg.	& 97.79\%  & 0.98  & 97.89\%  & 0.98  & 97.61\%  & 0.98  & 97.59\%  & 0.98 \\
            image-op          		& Linear reg.	& 86.48\%  & 0.86  & 83.00\%  & 0.83  & 86.36\%  & 0.86  & 86.76\%  & 0.87 \\
            neural-op         		& Linear reg.	& 98.91\%  & 0.99  & 98.76\%  & 0.99  & 98.65\%  & 0.99  & 98.81\%  & 0.99 \\
            image-op + neural-op 	& Linear reg.	& 92.44\%  & 0.92  & 91.03\%  & 0.91  & 92.35\%  & 0.92  & 92.67\%  & 0.93 \\
            \midrule
	        all               		& Quadratic reg.	& 94.44\%  & 0.94  & 93.55\%  & 0.94  & 94.92\%  & 0.95  & 94.95\%  & 0.95 \\
            cycle-op          		& Quadratic reg.	& 98.32\%  & 0.98  & 97.54\%  & 0.98  & 98.12\%  & 0.98  & 98.32\%  & 0.98 \\
            image-op          		& Quadratic reg.	& 88.54\%  & 0.88  & 85.21\%  & 0.85  & 88.52\%  & 0.88  & 88.95\%  & 0.89 \\
            neural-op         		& Quadratic reg.	& \highlightG{99.14\%}  & 0.99  & 98.45\%  & 0.98  & \highlightG{99.06\%}  & 0.99  & 98.98\%  & 0.99 \\
            image-op + neural-op 	& Quadratic reg.	& 93.18\%  & 0.93  & 92.26\%  & 0.92  & 93.84\%  & 0.94  & 93.95\%  & 0.94 \\
            \bottomrule
        \end{tabular}
    \end{subtable}
    \vspace{1em}
    \begin{subtable}{\textwidth}
        \centering
        \caption{Model performance results for mid-cut segmentation, applied to each phase type}
        \label{tab:model_performance_mid}
        \begin{tabular}{@{}llrrrrrrrr@{}}
            \toprule
            \multicolumn{1}{c}{\textbf{Phase type}} & 
            \multicolumn{1}{c}{\textbf{Signature}} & 
            \multicolumn{1}{c}{\textbf{BDT acc.}} & 
            \multicolumn{1}{c}{\textbf{BDT F1}} & 
            \multicolumn{1}{c}{\textbf{DT acc.}} &
            \multicolumn{1}{c}{\textbf{DT F1}} &
            \multicolumn{1}{c}{\textbf{ET acc.}} &
            \multicolumn{1}{c}{\textbf{ET F1}} &
            \multicolumn{1}{c}{\textbf{RF acc.}} &
            \multicolumn{1}{c}{\textbf{RF F1}} \\
            \midrule
            all						& Linear reg.	& 94.88\%  & 0.95  & 94.51\%  & 0.95  & 95.16\%  & 0.95  & 95.13\%  & 0.95 \\
            cycle-op          		& Linear reg.	& 98.53\%  & 0.99  & 98.45\%  & 0.98  & 98.37\%  & 0.98  & 98.48\%  & 0.98 \\
            image-op          		& Linear reg.	& 88.41\%  & 0.88  & 85.44\%  & 0.85  & 88.34\%  & 0.88  & 88.62\%  & 0.89 \\
            neural-op         		& Linear reg.	& \highlightG{99.14\%}  & 0.99  & \highlightG{99.16\%}  & 0.99  & 98.78\%  & 0.99  & 98.98\%  & 0.99 \\
            image-op + neural-op 	& Linear reg.	& 93.31\%  & 0.93  & 91.92\%  & 0.92  & 93.50\%  & 0.93  & 93.75\%  & 0.94 \\
            \midrule
	        all						& Quadratic reg.	& 95.14\%  & 0.95  & 94.60\%  & 0.95  & 96.06\%  & 0.96  & 95.98\%  & 0.96 \\
            cycle-op          		& Quadratic reg.	& \highlightY{99.11\%}  & 0.99  & 98.65\%  & 0.99  & 98.93\%  & 0.99  & \highlightY{99.01\%}  & 0.99 \\
            image-op          		& Quadratic reg.	& 89.48\%  & 0.89  & 87.30\%  & 0.87  & 90.17\%  & 0.90  & 90.04\%  & 0.90 \\
            neural-op         		& Quadratic reg.	& \highlightG{99.54\%}  & 1.00  & \highlightG{99.16\%}  & 0.99  & \highlightG{99.19\%}  & 0.99  & \highlightG{99.42\%}  & 0.99 \\
            image-op + neural-op 	& Quadratic reg.	& 94.03\%  & 0.94  & 92.71\%  & 0.93  & 94.74\%  & 0.95  & 94.66\%  & 0.95 \\
            \bottomrule
        \end{tabular}
    \end{subtable}
\end{table*}
\setcounter{table}{0} 
\begin{table*}[htbp]
    \centering
    \label{tab:model_performance_extension}
    \begin{subtable}{\textwidth}
        \centering
        \setcounter{subtable}{3} 
        \caption{Model performance results for end-cut segmentation, applied to each phase type}
        \label{tab:model_performance_end}
        \begin{tabular}{@{}llrrrrrrrr@{}}
            \toprule
            \multicolumn{1}{c}{\textbf{Phase type}} & 
            \multicolumn{1}{c}{\textbf{Signature}} & 
            \multicolumn{1}{c}{\textbf{BDT acc.}} & 
            \multicolumn{1}{c}{\textbf{BDT F1}} & 
            \multicolumn{1}{c}{\textbf{DT acc.}} &
            \multicolumn{1}{c}{\textbf{DT F1}} &
            \multicolumn{1}{c}{\textbf{ET acc.}} &
            \multicolumn{1}{c}{\textbf{ET F1}} &
            \multicolumn{1}{c}{\textbf{RF acc.}} &
            \multicolumn{1}{c}{\textbf{RF F1}} \\
            \midrule
            all						& Linear reg.	& 95.10\%  & 0.95  & 95.03\%  & 0.95  & 95.57\%  & 0.96  & 95.75\%  & 0.96 \\
            cycle-op				& Linear reg.	& 98.45\%  & 0.98  & 98.20\%  & 0.98  & 98.35\%  & 0.98  & 98.40\%  & 0.98 \\
            image-op				& Linear reg.	& 89.86\%  & 0.90  & 88.08\%  & 0.88  & 89.91\%  & 0.90  & 90.37\%  & 0.90 \\
            neural-op				& Linear reg.	& 98.76\%  & 0.99  & 98.53\%  & 0.99  & 98.37\%  & 0.98  & 98.60\%  & 0.99 \\
            image-op + neural-op	& Linear reg. 	& 93.75\%  & 0.94  & 93.13\%  & 0.93  & 94.11\%  & 0.94  & 94.27\%  & 0.94 \\
            \midrule
	        all						& Quadratic reg.	& 94.48\%  & 0.94  & 94.94\%  & 0.95  & 96.11\%  & 0.96  & 96.12\%  & 0.96 \\
            cycle-op				& Quadratic reg.	& 98.48\%  & 0.98  & 97.99\%  & 0.98  & 98.40\%  & 0.98  & 98.32\%  & 0.98 \\
            image-op				& Quadratic reg.	& 89.13\%  & 0.89  & 88.77\%  & 0.89  & 91.08\%  & 0.91  & 90.93\%  & 0.91 \\
            neural-op				& Quadratic reg.	& 98.81\%  & 0.99  & 98.60\%  & 0.99  & 98.50\%  & 0.98  & 98.63\%  & 0.99 \\
            image-op + neural-op	& Quadratic reg.	& 93.28\%  & 0.93  & 93.24\%  & 0.93  & 95.07\%  & 0.95  & 94.86\%  & 0.95 \\
            \bottomrule
        \end{tabular}
    \end{subtable}
\end{table*}
\stepcounter{table}  

Note that although prediction accuracies for different trainings from \Cref{tab:model_performance} are fairly close, differences are reliable. Calculating the Coefficient of Variation (CV)\footnote{$\text{CV} = \frac{\sigma}{\mu} \times 100$ \%} across folds reveals that per fold results are highly consistent. The observed CV values for highlighted trainings are 0.23\% or less, which is extremely low.

\subsection{Metrics to consider}
Data from different metrics result in different prediction performances, which is the motivation behind our focus on the data from the \emph{electrical current} metric. Selection of a metric beforehand cannot be directly deduced, but the effectiveness holds throughout. Therefore, it is a matter of trial. For brevity, the results for other metrics are not covered in \Cref{tab:model_performance}.

\subsection{Signature levels}
Signatures representing execution behaviour within arbitrary segments of data are calculated as regression functions. Higher orders of regression functions (quadratic, cubic, etc.) result in more accurate representation of data points and better prediction performance, but impose extra computational cost during data preprocessing. Depending on the scale and the deployment platform of choice, the difference in computational cost might be negligible or considerable. Restricting the signature generation to linear regression functions could be a legitimate priority in high-scale and resource-constrained set-ups, e.g., edge, embedded, and serverless computing deployments, due to computational and energy costs. The data processing workflow from \Cref{fig:data_processing} has to be applied for inference as well, for which, embedded or cloud-based serverless deployments are a common choice in the industry.

\subsection{Data segmentation effects}
The choice of data segmentation is the most influential aspect. In the absence of internal phase knowledge, the dataset generation workflow is limited to the one phase type representing operational cycles, \texttt{cycle-op}, i.e., full processing of an input instance (image). Looking at the top\footnote{We consider above 99.00\% accuracy as a top performance result.} classification performance results, BDT, ET and RF algorithms come forth for both \emph{full} and \emph{mid} data cuts. These values are indicated with yellow highlight. All top results belong to quadratic regression signatures. The very best result, 99.11\%, belongs to the combination of BDT, \texttt{cycle-op}, \emph{mid} data cut, and quadratic regression as signature. From \Cref{tab:model_performance}, it is clear that lower knowledge positions require more extra computational effort for better results.

Upon the presence of internal knowledge though, results improve for both \emph{full} and \emph{mid} data cuts, while using both linear and quadratic regression signatures. The consistent observation across the board in \Cref{tab:model_performance_full} points to the superior prediction performance from the \texttt{neural-op} phase type. This is a somewhat intuitive outcome, as the bulk of the computation happens during this phase. These values are indicated with green highlight. The absolute best result of all tables, 99.54\%, belongs to the combination of BDT, \texttt{neural-op}, \emph{mid} data cut, and quadratic regression as signature. If lower computational effort is mandated (linear regression), the BDT with above 99\% performance is still very suitable, in this case combined with the \emph{full} data cut.

\subsection{ML algorithm of choice}
The ML algorithm choices are already narrowed down to tree-based algorithms and these are very performant. Amongst these, BDT and RF have a consistent edge over DT and ET, with BDT posting the accuracy of 99.54\% (\Cref{tab:model_performance_mid}). An interesting observation is the higher performance of DT with linear regression under the \emph{full} data cut.

\subsection{Covered information positions}
While we have listed, implemented and applied techniques to manipulate data quality and achieve a data spectrum of different levels, we have not yet applied the traditional ML solution to these. It will be our immediate future work. As such, the data behind the results from \Cref{tab:model_performance} is from a rich data position.

\begin{figure}[htbp]
	\centering
	\includegraphics[width=0.7\linewidth]{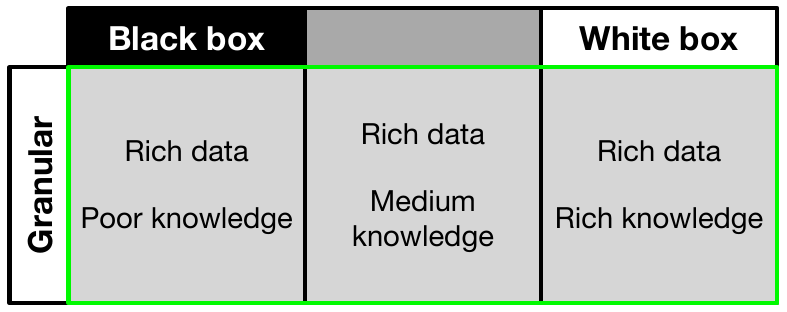}
	\caption{Considering the comprehensiveness of data and the variations in knowledge position in our cases, we are covering the bottom row of the information position quadrant.}
	\label{fig:quadrant_coverage}
\end{figure}

With the provided results and the information position quadrant concept, we can fill the cells on the bottom row, i.e., \Cref{fig:quadrant_coverage}. The knowledge dimension is clearly divided between white box and black box perspectives. Poor knowledge position is reflected by datasets built using the cycle information alone, while rich data assumes more granular phase awareness. The latter is represented by datasets built using more granular phase cuts, \texttt{image-op}, \texttt{neural-op}, and relevant combinations, i.e., \texttt{all} and \texttt{image-op + neural-op}.

\subsection{Limitations}
The InfoPos conceptual framework in its current form, is applicable to fault detection and identification solutions using machine learning. Additionally, the target is solely industrial CPS. Application of InfoPos to different types of solutions will require a one-time analysis, similar to what is described in this paper. At the same time, while close similarities exist between our demonstrator and production grade subsystems, the study shall be applied to a production use-case as well. This will be the immediate future work.

We do not compare to other solutions, as proof of superior performance is not the point here. The scope of this paper is to evaluate the implications of different information positions on solution effectiveness and ultimately aid with solution design.

\section{Related work}
\label{sec:related_work}
Experiments with different data positions are not covered in this paper. However, it is worth mentioning the two approaches towards data quality: source data quality and dataset quality. Source data refers to time-series data collected from physical or virtual sensors.
Source data is processed and turns into datasets for ML model training, which is the latter type of data. Publications considering the effects of data quality on ML algorithms~\cite{Mohammed:2024:EDQM, Foroni:2021:EEED, Frenay:2014:CPLN, Li:2021:CSEI, Neutatz:2022:DCAW, Shah:2024:HDCD}, unanimously focus on dataset quality, which can be defined with a number of dimensions itself~\cite{Mohammed:2024:EDQM}. It can be observed that publications delving into source data quality (many are listed in~\cite{Teh:2020:SDQR}, e.g., \cite{Zhang:2018:BNMD, Chen:2015:GBMD, Bosman:2015:EILD}), do so primarily in combination with traditional methods with no machine learning applied. We shall consider both source data quality and dataset quality in our future work. What stands out is the absence of knowledge variation and relevant effects. The closest concept to the consideration of knowledge as a separate dimension is \enquote{task-dependent quality}~\cite{Foroni:2021:EEED}, which still considers data quality in the context of the task it is being used for, i.e., a variable quality limit.

A similar, but relatively dated concept is knowledge-based configuration or configuration research~\cite{Stumptner:1997:OKBC}. The concept has been about selection of the right configuration for an end product, which has evolved into knowledge being an input to ML models. If an ML-assisted solution is consider as the product, theoretically it should be possible to have an AI agent design the most effective and most efficient solution. To the extent of our knowledge, there are no studies exploring combined effects of available knowledge and data qualities and utilising such information in streamlining the design process.

The particular anomaly identification solution examined here has shown its effectiveness and it is out of the scope of this paper to perform comparisons with other methods. In fact, there is a rich body of literature on anomaly and fault detection/identification solutions, using traditional techniques (for instance statistical methods)~\cite{Ibidunmoye:2015:PADB}, a combination of data processing and traditional ML~\cite{Ibidunmoye:2015:PADB, Odyurt:2022:IRIC}, or deep learning algorithms~\cite{Darban:2024:DLTS}. Finding an effective approach on streamlining the solution design effort is what is being sought in this paper.

\section{Conclusion and future work}
\label{sec:conclusion}
Presented exploratory results show that the InfoPos framework can definitively assist in selection of the combination of elements and configuration applied to fault identification. These include details such as the level of preprocessing, e.g., regression, selected data portion, and ML algorithm of choice. The possession of such awareness, upfront, will lead to a much more streamlined design search and implementation effort.

\paragraph*{Summary of results}
Considering the numerous available configuration, e.g., selection of uninformed data cut, selection of signature accuracy (regression degree), selection of operational phase (if available), and how expensive it is to experiment training ML models, the following are conclusions for this particular ML-assisted fault identification:
\begin{itemize}
	\item Application of uninformed segmentation and the choice of \emph{mid} data cuts is preferred, for both rich and poor knowledge positions.
	\item If linear regression signatures are mandated (computational and energy efficiency), \emph{full} data cuts work best.
	\item As expected, \emph{ini} and \emph{end} data cuts are not useful on their own, unless for very rare fault types contained in these.
	\item In the case of informed segmentation, combinations of phases (\texttt{all} and \texttt{image-op + neural-op}) do not lead to better results than using single phase types.
	\item BDT can be selected as the traditional ML algorithm of choice.
	\item Lack of internal operational knowledge could be compensated with higher degree regressions.
\end{itemize}

When it comes to the question of reusability, the conclusion holds for the type of fault identification solution evaluated in this paper, i.e., ML models trained with constructs (signatures in this case) based on data segmentation. While some of the findings are explainable with theory, turning them into general rules, other findings are empirical in nature. Examples of general findings are the effectiveness of \emph{mid} data cuts, or better results when regression functions of higher degree being used. Case-specific variables, such as the discovery of the most effective informed segmentation (\texttt{neural-op} for our use-case), or the type of ML algorithm, will need the execution of a minimal viable example as an analytic step. The industry utilising this type of CPS, e.g., semiconductor photolithography, production printing, even MRI machines in the health industry, is by no means small. Fault identification solutions are equally valuable across the board.

Immediate next steps for us are to complete the information position matrix with representative scenarios of varying data quality, using data degradation techniques listed in \Cref{subsec:simulating_data_positions}, as well as execution of diverse types of ML-assisted solutions. The latter will include Deep Neural Networks and Transformer-based alternative designs.

\begin{acks}
This publication is part of the project ZORRO with project number KICH1.ST02.21.003 of the research programme Key Enabling Technologies (KIC), which is (partly) financed by the Dutch Research Council (NWO).
\end{acks}
